\documentclass[10pt,twocolumn,letterpaper]{article}

\usepackage{formattings/hadianeh}
\usepackage{formattings/macros}

\usepackage{tikz}

\usepackage{placeins}

\usepackage{formattings/iccv}
\usepackage{times}
\usepackage{epsfig}
\usepackage{graphicx}
\usepackage{amsmath}
\usepackage{amssymb}

\usepackage{booktabs}

  \usepackage{caption}
  \usepackage{subfig}

\usepackage{algorithm}
\usepackage{algorithm2e}
\usepackage{algorithmic}

\usepackage{calrsfs}
\newcommand{\La}{\mathcal{L}}
\DeclareMathAlphabet{\pazocal}{OMS}{zplm}{m}{n}

\usepackage{caption}
\usepackage[font={sf,footnotesize,bf}]{caption}
\usepackage[labelfont={bf,it},textfont=it]{caption}

\usepackage[pagebackref=true,breaklinks=true,letterpaper=true,colorlinks,bookmarks=false]{hyperref}
\hypersetup{
	colorlinks=true,
	linkcolor=blue,          
    citecolor=blue,        
    filecolor=blue,      
    urlcolor=blue
}
 \iccvfinalcopy 

\def\iccvPaperID{6541\vspace{-4ex}} 

\ificcvfinal\fi

\usepackage[subtle]{savetrees}

\newcommand{\sinreq}[0]{\textsf{\small SinReQ}\xspace}

\begin{document}

\title{SinReQ: Generalized Sinusoidal Regularization for Low-Bitwidth Deep Quantized Training\vspace{-2ex}}

\author{Ahmed T. Elthakeb \quad Prannoy Pilligundla \quad Hadi Esmaeilzadeh \\
\textbf{A}lternative \textbf{C}omputing \textbf{T}echnologies ({\color[HTML]{0B6121}{\textbf{ACT}}}) Lab\\
University of California San Diego \\
{\sffamily\small {\{\href{mailto:a1yousse@eng.ucsd.edu}{a1yousse},\ \href{mailto:ppilligu@eng.ucsd.edu}{ppilligu}, 
\href{mailto:hadi@eng.ucsd.edu}{hadi}\}@eng.ucsd.edu}}}

{ 
}

\maketitle

\begin{abstract}
Deep quantization of neural networks (below eight bits) offers significant promise in reducing their compute and storage cost.
Albeit alluring, without special techniques for training and optimization, deep quantization results in significant accuracy loss.
To further mitigate this loss, we propose a novel sinusoidal regularization, called \sinreq \footnote{\textit{Accepted and presented in ICML 2019 Workshop on Understanding and Improving Generalization in Deep Learning, Long Beach, California, 2019. Copyright 2019 by the author(s).}}, for deep quantized training.
\sinreq adds a \textit{periodic} term to the original objective function of the underlying training algorithm.
\sinreq exploits the periodicity, \emph{differentiability}, and the desired convexity profile in sinusoidal functions to automatically propel weights towards values that are inherently closer to quantization levels.
Since, this technique does not require invasive changes to the training procedure, \sinreq can harmoniously enhance quantized training algorithms.
\sinreq offers generality and flexibility as it is not limited to a certain bitwidth or a uniform assignment of bitwidths across layers.
We carry out experimentation using the AlexNet, CIFAR-10, ResNet-18, ResNet-20, SVHN, and VGG-11 DNNs with three to five bits for quantization and show the versatility of \sinreq in enhancing multiple quantized training algorithms, DoReFa~\cite{Zhou2016DoReFaNetTL} and WRPN~\cite{Mishra2017WRPNWR}.
Averaging across all the bit configurations shows that \sinreq closes the accuracy gap between these two techniques and the full-precision runs by 32.4\% and 27.5\%, respectively.
That is improving the absolute accuracy of DoReFa and WRPN by 2.8\% and 2.1\%, respectively.

%
%
%
\end{abstract}


\section{Introduction}
\label{sec:intro}

%
%
Despite the success of DNNs in various domains~\cite{DBLP:journals/nature/LeCunBH15,sirius,bpzip,DBLP:conf/nips/KrizhevskySH12}, their compute efficiency hinders effective deployment in resource-limited platforms such as embedded sensors or mobile devices ~\cite{DBLP:journals/pieee/SzeCYE17}.
Quantization, in general, and deep quantization, in particular, aim to not only reduce the compute requirements of DNNs but also significantly reduce their memory footprint~\cite{Zhou2016DoReFaNetTL,Mishra2017WRPNWR,Hubara2017QNN,bitfusion,Judd2016StripesBD}.
Nevertheless, without specialized training and optimization  training algorithms, quantization can diminish the accuracy.
As such, several techniques have been proposed that aim to train DNNs in quantized mode with as low as possible loss in accuracy~\cite{DBLP:conf/nips/CourbariauxBD15,DBLP:conf/icml/GuptaAGN15,DBLP:journals/jmlr/HubaraCSEB17,DBLP:conf/iclr/ZhouYGXC17}.
However, eliminating the loss has proven to be illusive.

This paper aims to provide a new mechanism that enhances these techniques and significantly closes the remaining gap between deeply quantized and full precision networks.
As such, we propose a sinusoidal regularization technique, a differentiable term, that naturally pushes the weight values toward the quantization levels. 
Sinusoidal functions have inherent periodic minima which can be exploited to push the weights towards the desired quantization levels.
Thus, by adding a sinusoidal regularizer to the original objective function, our method automatically yields quantization friendly weights for any given bitwidths.
In fact, the original optimization procedure itself is harnessed for this purpose, which is enabled by the differentiability of the sinusoidal regularization term.
As such, quantized training algorithms~\cite{Zhou2016DoReFaNetTL,Mishra2017WRPNWR} that still use some form of backpropagation~\cite{rumelhart:errorpropnonote} can effectively utilize the proposed mechanism.
Consequently, this technique, called \sinreq, can potentially enhance quantized training algorithms by modifying their loss.

\sinreq offers generality and can be used with different bitwidths by setting the periodicity of the regularizer according to the desired bitwidth.
Moreover, the proposed technique is flexible and a dedicated sinusoidal term for each layer with different periods can enable heterogenous quantization across the layers.
The \sinreq regularization can also be applied for training a model from scratch, or for fine-tuning a pretrained model.

We evaluate \sinreq using different bitwidth assignments across for different DNNs (CIFAR-10, ResNet-20, SVHN, and VGG-11).
To show the versatility of \sinreq, it is used with two different quantized training algorithms, DoReFa~\cite{Zhou2016DoReFaNetTL} and WRPN~\cite{Mishra2017WRPNWR}.
Over all the bitwidth assignments, the proposed regularization, on average, improves the top-1 accuracy of DoReFa and WRPN by 2.8\% and 2.1\%, respectively.
That is, closing the gap between the quantized network and a full-precision netwrok by 37.1\% in the case of DoReFa~\cite{Zhou2016DoReFaNetTL} and 36.5\% in the case of WRPN~\cite{Mishra2017WRPNWR}.

\section{Related Work}

\sinreq, which is a regularization technique, is complementary to the previously proposed quantized training~\cite{DBLP:conf/nips/CourbariauxBD15,DBLP:conf/icml/GuptaAGN15,DBLP:journals/jmlr/HubaraCSEB17,DBLP:conf/iclr/ZhouYGXC17} and binarization~\cite{NIPS2016BNN,Rastegari2016XNORNetIC} algorithms and can potentially augment their training procedure.
Additionally, there is a line of research that aims to incorporate the distance between the quantized levels and the full-precision weights in the training loss~\cite{DBLP:journals/corr/abs-1808-05779,DBLP:conf/iclr/HouYK17,DBLP:conf/iclr/HouK18,DBLP:journals/corr/abs-1809-00095,DBLP:conf/eccv/ZhangYYH18}.
In contrast, \sinreq, utilizes the periodic nature of sinusoidal function to push the weight values to the quantization levels possible by the allocated bitwidth to each layer.
%

\niparagraph{Training algorithms for quantized neural networks.}
There have been several techniques \cite{Zhou2016DoReFaNetTL,Zhu2016TrainedTQ,Mishra2017WRPNWR} that train a neural network in a quantized domain after the bitwidth of the layers is determined manually.
DoReFa-Net \cite{Zhou2016DoReFaNetTL} quantizes weights, activations and gradients of neural networks using different bitwidths. They suggest maintaining a high-precision floating point copy of the weights while feeding quantized weights into backprop.
WRPN \cite{Mishra2017WRPNWR} introduces a scheme to train networks from scratch using reduced-precision activations by decreasing the precision of both activations and weights and increasing the number of filter maps in a layer. 
\cite{Zhu2016TrainedTQ} performs the training phase of the network in full precision, but for inference uses ternary weight assignments. 
For this assignment, the weights are quantized using two scaling factors which are learned during training phase.
PACT~\cite{Choi2018PACTPC} introduces a quantization scheme for activations, where the variable $\alpha$ is the clipping level and is determined through a gradient descent based method. More recently, VNQ~\cite{DBLP:conf/iclr/AchterholdKSG18} uses a variational Bayesian approach for quantizing neural network weights during training. 

\sinreq is a complimentary method that can potentially enhance these algorithms.
The paper demonstrates this feature concretely in the context of DoReFa~\cite{Zhou2016DoReFaNetTL} and WRPN~\cite{Mishra2017WRPNWR} training algorithms.


\niparagraph{Binarized and ternarized neural networks.}
Extensive work, \cite{Hubara2017QNN,Rastegari2016XNORNetIC,Li2016TernaryWN} focuses on binarized neural networks, which impose accuracy loss but reduce the bitwidth to lowest possible level.
In BinaryNet \cite{DBLP:conf/nips/HubaraCSEB16}, an extreme case, a method is proposed for training binarized neural networks which reduce memory size, accesses and computation intensity at the cost of accuracy.
XNOR-Net \cite{Rastegari2016XNORNetIC} leverages binary operations (such as XNOR) to approximate convolution in binarized neural networks. 
BinaryNet and XNOR-Net alleviate most of the multiplications needed in the forward and backward passes of the network and thereby achieve high performance in comparison to full-precision networks.
On the other side, in a weight ternarized network, zero is used as an additional quantized value. \cite{Li2016TernaryWN} introduces ternary-weight networks, in which the weights are quantized to {-1, 0, +1} values by minimizing the Euclidian distance between full-precision weights and their ternary assigned values.
In~\cite{DBLP:conf/iclr/ZhuHMD17} different scaling factors are introduced to the ternarized weights. The scaling parameters are learned by gradient descent.

None of these techniques propose sinusoidal regularization to make the weight values more quantization friendly as training progresses.
%

\niparagraph{Loss-aware weight quantization.}
Recent works pursued loss-aware minimization approaches for quantization. 
\cite{DBLP:conf/iclr/HouYK17,DBLP:conf/iclr/HouK18} developed approximate solutions using proximal Newton algorithm to minimize the loss function directly under the constraints of low bitwidth weights.
\cite{DBLP:journals/corr/abs-1809-00095} proposed to learn the quantization of DNNs through regularization by introducing a learnable regularization coefficient to find low bitwidth models efficiently in training. 
\cite{DBLP:conf/eccv/ZhangYYH18} proposed an adaptive technique to jointly train a quantized, bit-operation-compatible DNN and its associated quantizers, as opposed to using fixed, handcrafted quantization schemes such as uniform or logarithmic quantization.

Although these techniques use regularization to guide the process of quantized training, they do not explore the use of periodic differentiable trigonometric functions.

%
%
In the context of using periodic functions for regularization, there is a concurrent work \cite{DBLP:journals/corr/abs-1811-09862} which was published on ArXiv around two weeks prior to the first publication of this work in \cite{releq_v1}. In addition to what they propose, we provide a rigorous analysis from convergence and optimization points of view and a comprehensive evaluation across various DNNs with comparative studies to existing state-of-the-art techniques.

\SetKwComment{Comment}{$\triangleright$\ }{}

\section{Sinusoidal Regularization for Automatic Quantization during Training}
\label{sec:method}
%
Our proposed method SinReQ exploits weight regularization in order to automatically quantize a neural network while training. To that end, Sections~\ref{sec:loss} to \ref{sec:weight_decay} describe the role of regularization in neural networks and then Section~\ref{sec:sinreq} explains SinReQ in more detail.
%
%

\subsection{Loss Landscape of Neural Networks}
\label{sec:loss}
Neural networks' loss functions are known to be highly non-convex and generally very poorly understood. It has been empirically verified that loss surfaces for large neural networks have many local minima that are essentially equivalent in terms of test error \cite{DBLP:journals/corr/ChoromanskaHMAL14}, \cite{DBLP:journals/corr/abs-1712-09913}. 
Moreover, converging to one of the many good local minima proves to be more useful as compared to struggling to find the global minimum of the accuracy loss on the training set (which often leads to overfitting). 

This opens up and encourages a possibility of adding extra custom objectives to optimize for during the training process, in addition to the original objective (i.e., to minimize the accuracy loss). The added custom objective could be with the purpose of increasing generalization performance or imposing some preference on the weights values.
Regularization is one of the major techniques that makes use of such facts as discussed in the following subsection.
%
\begin{figure*}
  \centering 
  \includegraphics[width=0.8\textwidth]{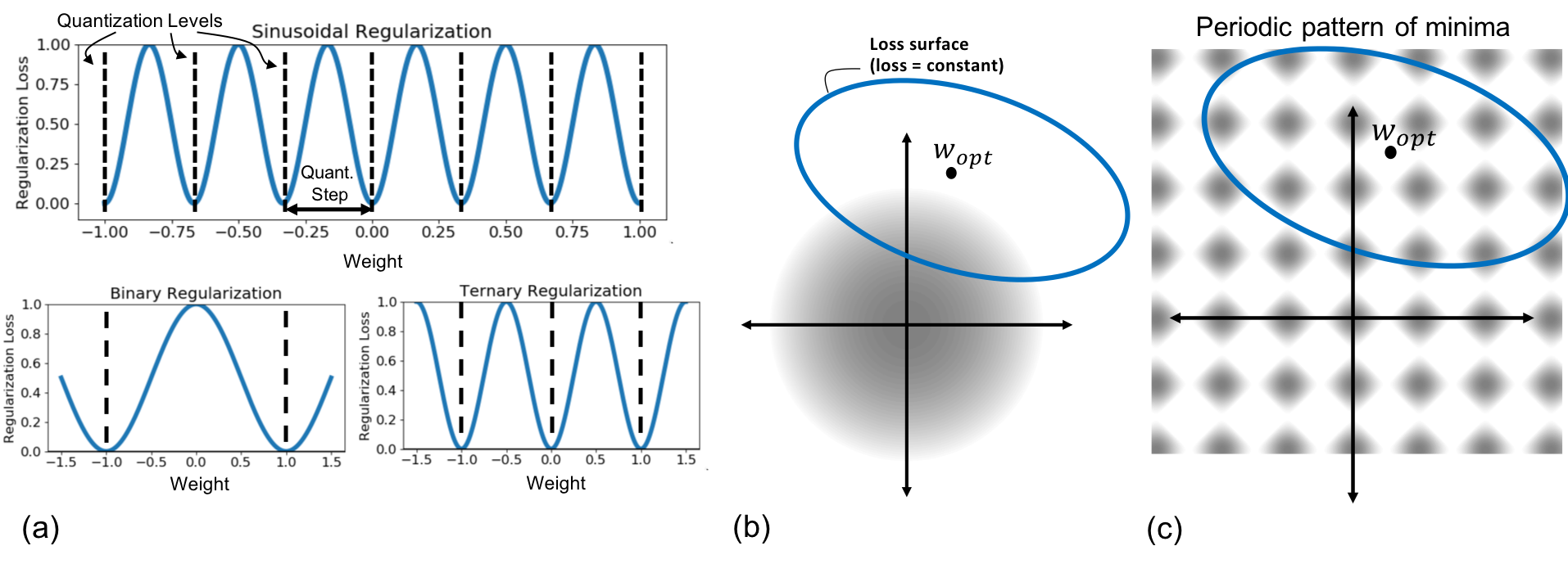}

  \caption{(a) Generalized SinReQ profile adapting for arbitrary bitwidths including binary and ternary quantization. (b) and (c) depict a geometrical sketch for a hypothetical loss surface (original objective function to be minimized) and an extra regularization term in 2-D weight space for weight decay and SinReQ respectively. $w_{opt}$ is the optimal point just for the loss function alone. }
  \label{fig:overview}
  \vspace{-0.5cm}
\end{figure*}
\subsection{Regularization in Neural Networks}
\label{sec:regularization}
Neural networks often suffer from redundancy of parameterization and consequently they commonly tend to overfit.
Regularization is one of the commonly used techniques to enhance generalization performance of neural networks.
Regularization effectively constrains weight parameters by adding a term (regularizer) to the objective function that captures the desired constraint in a soft way. This is achieved by imposing some sort of preference on weight updates during the optimization process. As a result, regularization seamlessly leads to unconditionally constrained optimization problem instead of explicitly constrained which, in most cases, is much more difficult to solve.

\textit{Weight decay}, which is the conventional regularizer used while training neural networks, is described next.
%

\subsection{Conventional Regularization: Weight Decay}
\label{sec:weight_decay}
The most commonly used regularization technique is known as \textit{weight decay}, which aims to reduce the network complexity by limiting the growth of the weights. It is realized by adding a term to the objective function that penalizes large weight values 
 \begin{equation}
E(w) = E_o(w) + \frac{1}{2} \lambda \sum_{i}\sum_{j} {w_{ij}}^2 
 \end{equation}
where $E_o$ is the original loss measure, and $\lambda$ is a parameter governing how strongly large weights are penalized. 
$w_{ij}$ denotes the $j$-th element of the vector $w_i$ and summation is over all layers in the network.

\begin{algorithm}
\caption{SinReQ regularization for training neural networks} 
\textbf{Input:} Full-precision weights $\{\overline{w}_{l}\}$ for each layer $l=1:L$; where $L$ is the total number of layers, $qbit_{l}$ quantization bits, $\lambda_{ql}$ regularization strength (optionally per layer)
\begin{algorithmic} [1] 
\STATE {\textcolor{gray}{\COMMENT {{Forward Propagation}}}}
\FOR{$i = \ 1$ to $L$} 
\STATE $step_i = (2^{qbit_i}-0.5)$ , $\Delta_i = step_i/2$  \ \ \textcolor{gray}{\COMMENT {DoReFa}}
\STATE $step_i = (2^{qbit_i}-1.0)$ , $\Delta_i = 0$  \ \  \textcolor{gray}{\COMMENT{WRPN}}
\STATE $\La_{SinReQ, i} \ = \ mean(sin^2(\pi\times(\overline{w}_{i}+\Delta_i)/step_i))$
\STATE $\La_{SinReQ} \ = \ \La_{SinReQ} \ + \ \La_{SinReQ, i}$
\ENDFOR
\STATE Compute the overall loss as the original loss ($\La_0$), and \sinreq loss: $\La=\La_o+ \lambda_{ql} \times \La_{SinReQ}$
\STATE \textcolor{gray}{\COMMENT{Backward Propagation}}
\FOR{$i = \ 1$ to $L$} 
\STATE Compute gradients $\nabla_i \La_o(\overline{w})$ and $\nabla_i \La_{SinReQ}(\overline{w})$
\STATE Update full-precision weights $\{\overline{w}_{i}\}$
\STATE Update regularization strength ($\lambda_{ql}$)
\ENDFOR
\end{algorithmic}
\label{alg:overvall_technique}
\end{algorithm}
\vspace{-0.3em}


%


\subsection{Periodic Regularization: SinReQ}
\label{sec:sinreq}
{\setlength{\parindent}{0cm}In this work, we propose a new type of regularization that is friendly to quantization. 
The proposed regularization is based on a periodic function (sinusoidal) that provides a smooth and differentiable loss to the original objective, Figure \ref{fig:overview} (a). 
The periodic regularizer has a periodic pattern of minima that correspond to the desired quantization levels. Such correspondence is achieved by matching the period to the quantization step based on a particular number of bits for a given layer.}
 \begin{equation}
E(w) = E_o(w) + \frac{1}{2} \lambda \sum_{i}\sum_{j} {w_{ij}}^2 + \frac{1}{2} \lambda_{q} \sum_{i}\sum_{j} {\sin^{2}\left({\frac {\pi w_{ij}} {2^{-qbits}-1}}\right)} 
 \end{equation}
  where $E_o$ is the original loss measure, and $\lambda_q$ is regularization strength that is a parameter governing how strongly weight quantization errors are penalized. 
  $w_ij$ denotes the $j$-th element of the vector $w_i$.
%
For the sake of simplicity and clarity, Figure \ref{fig:overview} (b) and (c) depict a geometrical sketch for a hypothetical loss surface (original objective function to be minimized) and an extra regularization term in 2-D weight space. 
For weight decay regularization, in Figure \ref{fig:overview} (b), the faded circular contours show that as we get closer to the origin, the regularization loss is minimized. $w_{opt}$ is the optimum just for the loss function alone and the overall optimum solution is achieved by striking a balance between the original loss term and the regularization loss term.

\vspace{0.5cm}
In a similar vein, Figure \ref{fig:overview} (c) shows a representation of the proposed regularization. A periodic pattern of minima pockets are seen surrounding the original optimum point. 
The objective of the optimization problem is to find the best solution that is the closest to one of those minima pockets where weight values are nearly matching the desired quantization levels, hence the name quantization-friendly.
Algorithm 1 details the implementation procedure of \sinreq regularization using LeNet as an example.
\subsection{Quantization Techniques} 
%
%
\niparagraph{DoReFa quantization.}
For k-bit representation with $k > 1$, the following function is proposed in~\cite{Zhou2016DoReFaNetTL} for weight quantization:
\vspace{0.3cm}
\begin{equation}
w_{q} = 2.quantize_k(\frac{tanh(w_f)}{2max(|tanh(w_f)|)} + \frac{1}{2}) - 1
\end{equation}
\vspace{0.3cm}
where $w_{q}$ and $w_{f}$ denote quantized weights and full precision weights respectively, and
\begin{equation}
quantize_k(x) = \frac{1}{2^k-1}round((2^k-1)x)
\end{equation}

$tanh$ limits the value range of weights $w_f$ to $\lbrack -1,1\rbrack$ before quantizing to k-bit, and $(\frac{tanh(w_f)}{2max(|tanh(w_f)|)} + \frac{1}{2})$ is a number in $\lbrack 0,1\rbrack$. 
$quantize_k$ will then quantize this number to k-bit fixed-point ranging in [0, 1].

\vspace{0.5cm}
\niparagraph{WRPN quantization.}
As proposed in~\cite{Mishra2017WRPNWR}, weights are first scaled and clipped to the $(-1.0, 1.0)$ range and quantized as per the following equation. 
\begin{equation}
w_{q} = \frac{round((2^{k-1} - 1)w_{f})}{2^{k-1} - 1}
\end{equation}
$k$ is the bitwidth used for quantization out of which $k-1$ bits are used for value quantization and one bit is used for sign. 
\vspace{0.5cm} \\
As it is shown in the above equations, different quantization techniques yield different quantization values. For example, quantized value of 0 is not used in DoReFa, which is not the case for WRPN. 
Hence, for each quantization technique, SinReQ loss minima can be seamlessly adjusted to the respective quantization values.
%

\SetKwComment{Comment}{$\triangleright$\ }{}
\begin{figure*}[ht]
  \centering
  \includegraphics[width=0.7\textwidth]{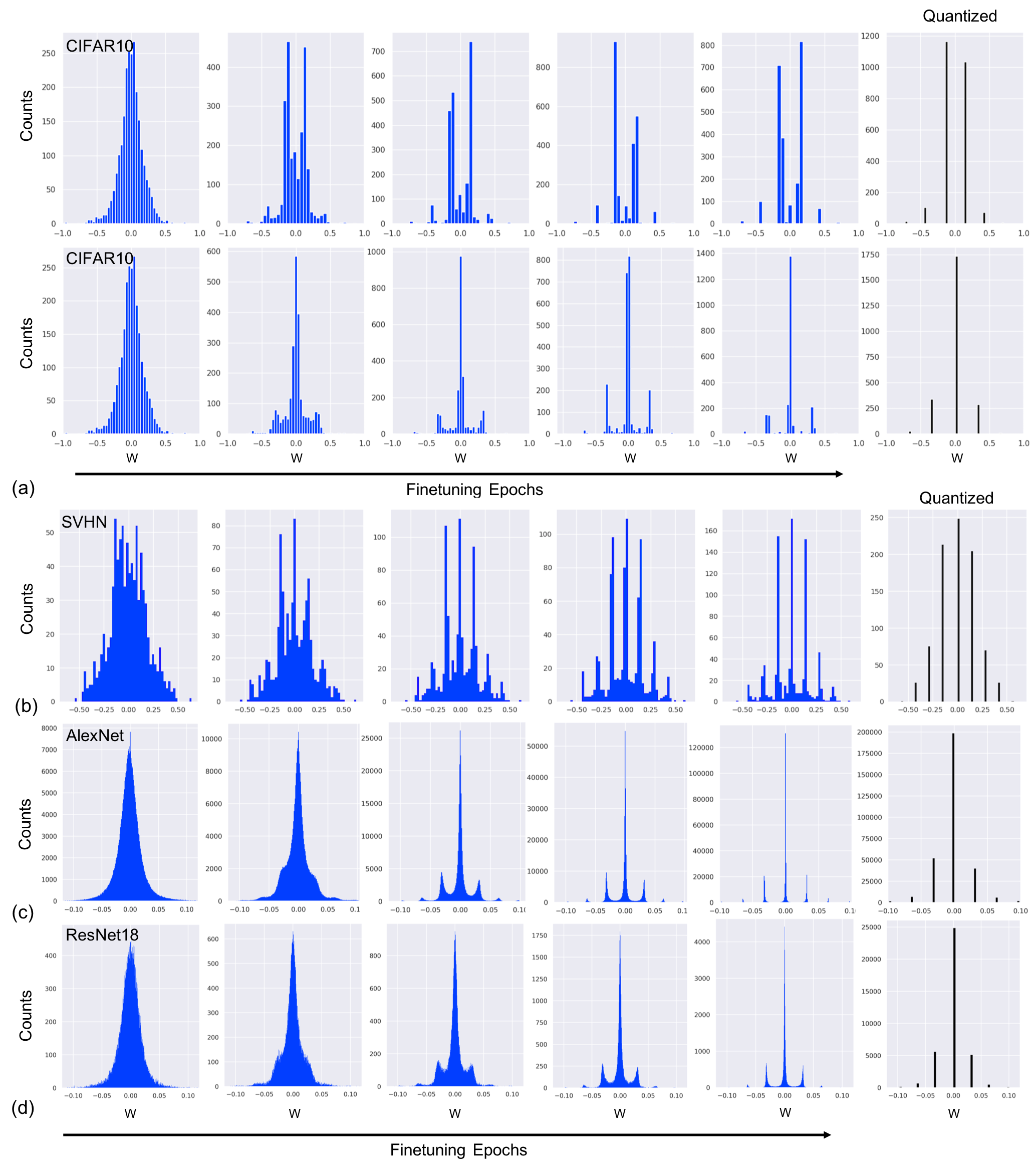}
  \caption{Evolution of weight distributions over training epochs (with the proposed regularization) at different layers and bitwidths for CIFAR10 and SVHN. (a) CIFAR10, second convolution layer with 3 bits, top row: mid-rise type of quantization (shifting by half a step to exclude zero as a quantization level); bottom row: mid-tread type of quantization (zero is included as a quantization level). (b) SVHN, top row: first convolution layer with 4 bits quantization. (c) AlexNet, second convolution layer with 5 bits quantization, and (d) ResNet-18, , second convolution layer with 5 bits quantization.}
  \label{fig:q_w_dist}
\end{figure*}
\begin{table*}
	\centering
	\caption{Summary of results comparing state-of-the-art methods DoReFa, and WRPN with and without SinReQ for different neural networks}	
	\includegraphics[width=0.55\linewidth]{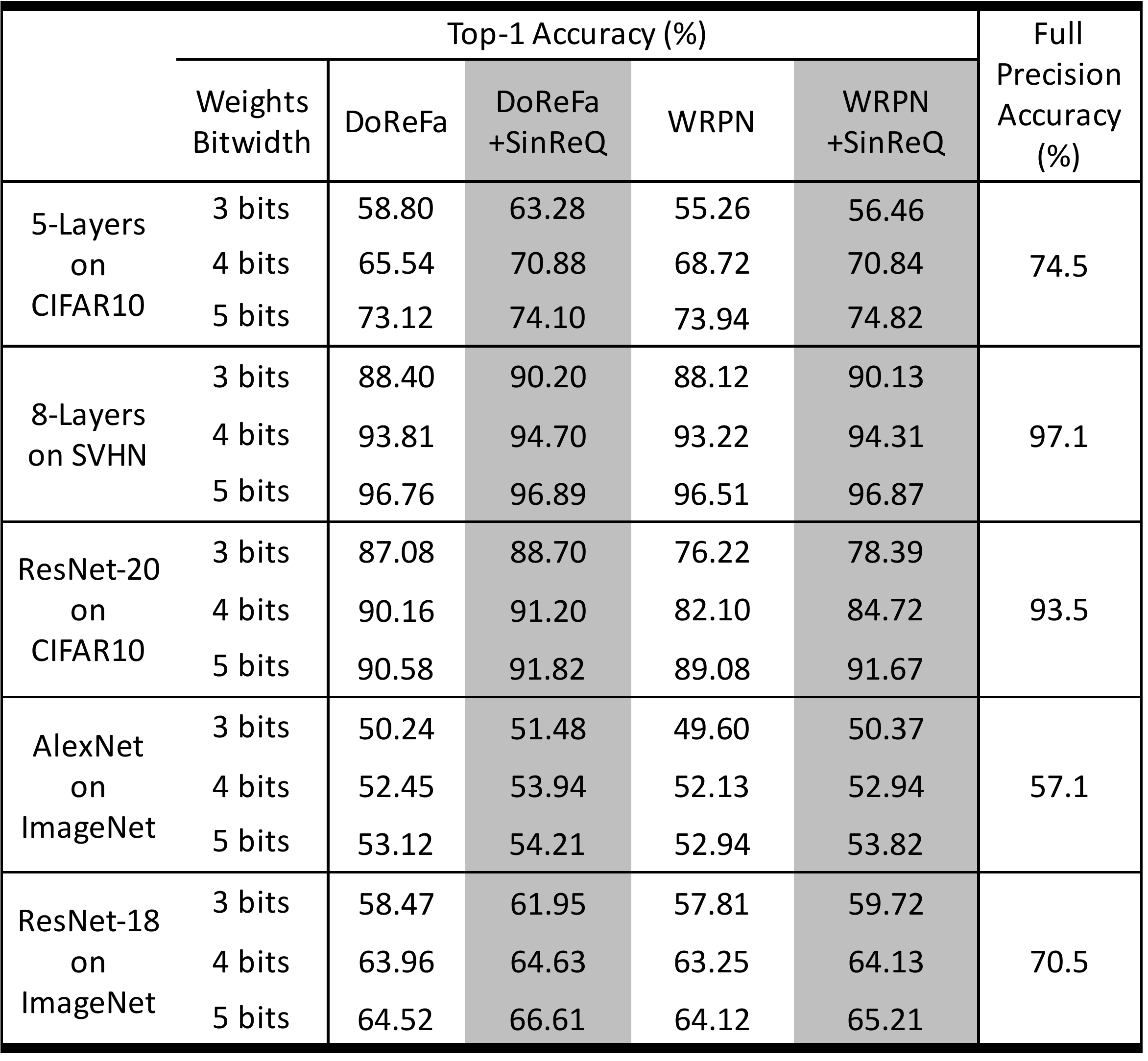}
	\label{table:results_summary}
\end{table*}

\section{Evaluation: SinReQ in Action}
To demonstrate the effectiveness of our proposed sinusoidal regularization, we evaluated it on three neural networks (CIFAR10, SVHN, and ResNet-20) with two image classification datasets (CIFAR10 and SVHN). 
Here, we focus on fine-tuning from a pretrained models as compared to training from scratch.
In the following subsections, we first start by introducing our experimental setup. Then, we analyze the impact of adding the proposed regularization to the original objective, during training, on the distribution of weights. 
Next, we highlight the main characteristics of the proposed method in terms of generalization for arbitrary-bitwidth quantization and customization for layer-wise optimization. 
Next, we compare to other existing methods by showing the achieved improvements across different benchmarks.
Lastly, we analyze the convergence behavior by introducing an example for training from scratch.

\niparagraph{Experimental Setup.}
We implemented our technique inside Distiller~\cite{neta_zmora_2018_1297430}, an open source framework for compression by Intel Nervana.
The reported accuracies for DoReFa and WRPN are with the built-in implementations in Distiller, which may not exactly match the accuracies reported in their respective papers.
However, an independent implementation from a major company provides an unbiased foundation for the comparisons.

\niparagraph{Semi-quantized weight distributions.}
Figure \ref{fig:q_w_dist} shows the evolution of weights distributions over fine-tuning epochs for different layers of CIFAR10 and SVHN networks. The high-precision weights form clusters and gradually converge around the quantization centroids as regularization loss is minimized along with the main accuracy loss.
The rate of convergence to the target quantization levels depends on (i) the number of fine-tuning epochs, (2) the regularization strength ($\lambda_q$). It is worth noting that $\lambda_q$ is a hyper-parameter that controls the tradeoff between the accuracy loss and the regularization loss. Fixed value can be presumed ahead of training or fine-tuning, however careful setting of such parameter can yield optimum results. \cite{DBLP:journals/corr/abs-1809-00095} considers the regularization coefficient as a learnable parameter.
%

\niparagraph{Arbitrary-bitwidth quantization.}
Considering the following sinusoidal regularizer, with $step_q$ denoting the quantization step, and $\Delta$ is an offset.
 \begin{equation}\label{eq:reg}
R(W) = \lambda_{q} \sum\limits_{i} {\sin^{2}\left({\frac {\pi w_i + \Delta} {step_q}}\right)} 
 \end{equation}
SinReQ provides generality in two aspects. First, the flexibility to adapt for arbitrary number of bits. 
The parameter $step_q$ controls the periodicity of the sinusoidal function. Thus, for any arbitrary bitwidth ($qbits$), $step_q$ can be tuned to match the respective quantization step. For uniform quantization:
 \begin{equation} 
step_q = {2^{-qbits}-1} 
 \end{equation}
Figure \ref{fig:q_w_dist} shows different examples of automatic gradual quantization of weights distributions at different bitwidths (3, 4, and 5 bits).

%
The second aspect of generality is the seamless accommodation for different quantization styles. There are two styles of uniform quantization: mid-tread and mid-rise. In mid-tread, zero is considered as a quantization level, while in mid-rise, quantization levels are shifted by half a step such that zero is not included as a quantization level. Ternary quantization, using $\{-1,0,1\}$, is an example of the former, while binary quantization is an example of the latter where two levels are used $\{-1,1\}$. 
Figure \ref{fig:q_w_dist} (a) shows the second convolution layer of CIFAR10 at 3 bits, top row: mid-rise type of quantization, and bottom row: mid-tread type of quantization.
%


\niparagraph{Layer-wise optimization.}
As different layers have different levels of sensitivity to the quantization bitwidth~\cite{DBLP:journals/corr/abs-1811-01704}, enabling layer-wise quantization opens the possibility for heterogenous bitwidth quantization and consequently more optimized quantized networks.
This can be achieved by adding a custom regularizer (as shown in equation \ref{eq:reg}) for each layer and sum over all layers. Then, we add the regularization losses of all layers to the main accuracy loss and pass the entire collective loss to the gradient-descent optimizer.

\begin{table}
	\centering
	\caption{Comparison with VNQ results for LeNet and DenseNet considering ternary weight quantization.}	
	\includegraphics[width=0.8\linewidth]{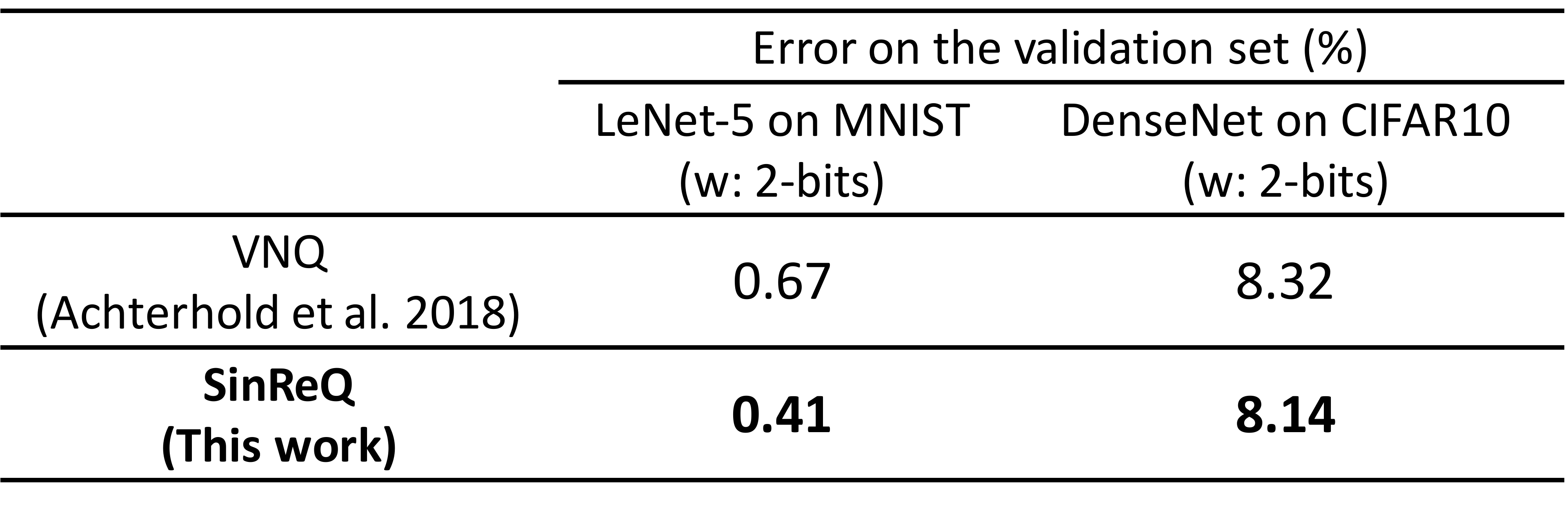}
	\label{table:VNQ}
	\vspace{-0.2ex}
\end{table}
\niparagraph{Comparison to existing methods.}
We evaluate our proposed approach with three different networks on two image classification datasets (CIFAR10 and SVHN).
We assess the efficacy of SinReQ on boosting the performance of existing methods for training quantized networks, DoReFa~\cite{Zhou2016DoReFaNetTL}, and WRPN~\cite{Mishra2017WRPNWR}.
Table \ref{table:results_summary} 
summarizes the accuracies obtained by DoReFa, and WRPN with and without SinReQ.
%
%
Results show that integrating SinReQ within the training algorithm achieves 2.8\%, and 2.1\% accuracy improvements on average to DoReFa, and WRPN methods respectively.

%
It is worth noting that, our proposed sinusoidal regularization can be considered as an auxiliary utility to existing methods to boost the efficiency of quantized training rather than being an alternative. As summarized in Table \ref{table:results_summary}, invoking SinReQ during training consistently yields improved accuracies as compared to the plain implementation of the considered quantized training methods. 

More recently, VNQ~\cite{DBLP:conf/iclr/AchterholdKSG18} uses a variational Bayesian approach for quantizing neural network weights during training. 
We, on the other hand, provide a regularization mechanism that works in tandem with other quantized training algorithms.
VNQ uses LeNet-5, and DenseNet for results. As table~\ref{table:VNQ} shows, for LeNet and DenseNet, SinReQ achieves 0.41\% and 8.14\% accuracy loss while the best result with VNQ are 0.67\% and 8.32\% respectively. 
%
%
\begin{figure}
\centering
  \includegraphics[width=0.4\textwidth]{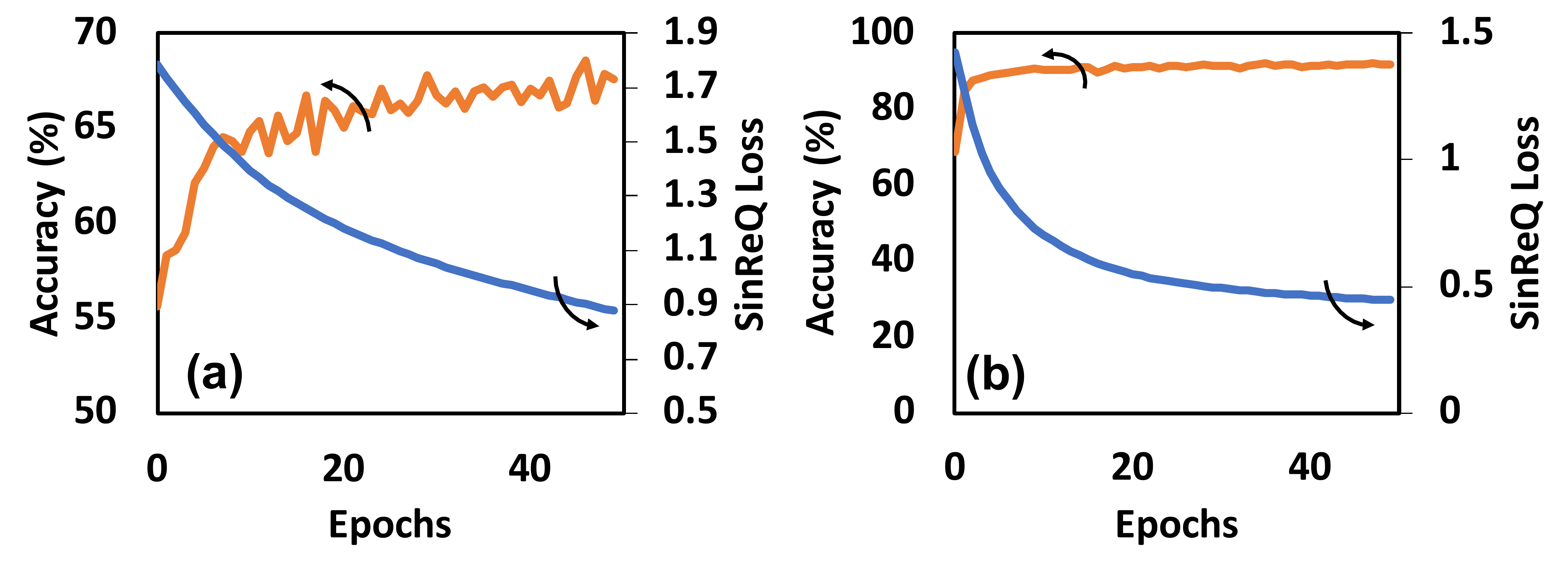}
  \caption{Convergence behavior: accuracy and \sinreq regularization loss over fine-tuning epochs for (a) CIFAR10, (b) SVHN }
  \label{fig:finetune}
\end{figure}
\begin{figure}
\centering
  \vspace{-12pt}
  \includegraphics[width=0.4\textwidth]{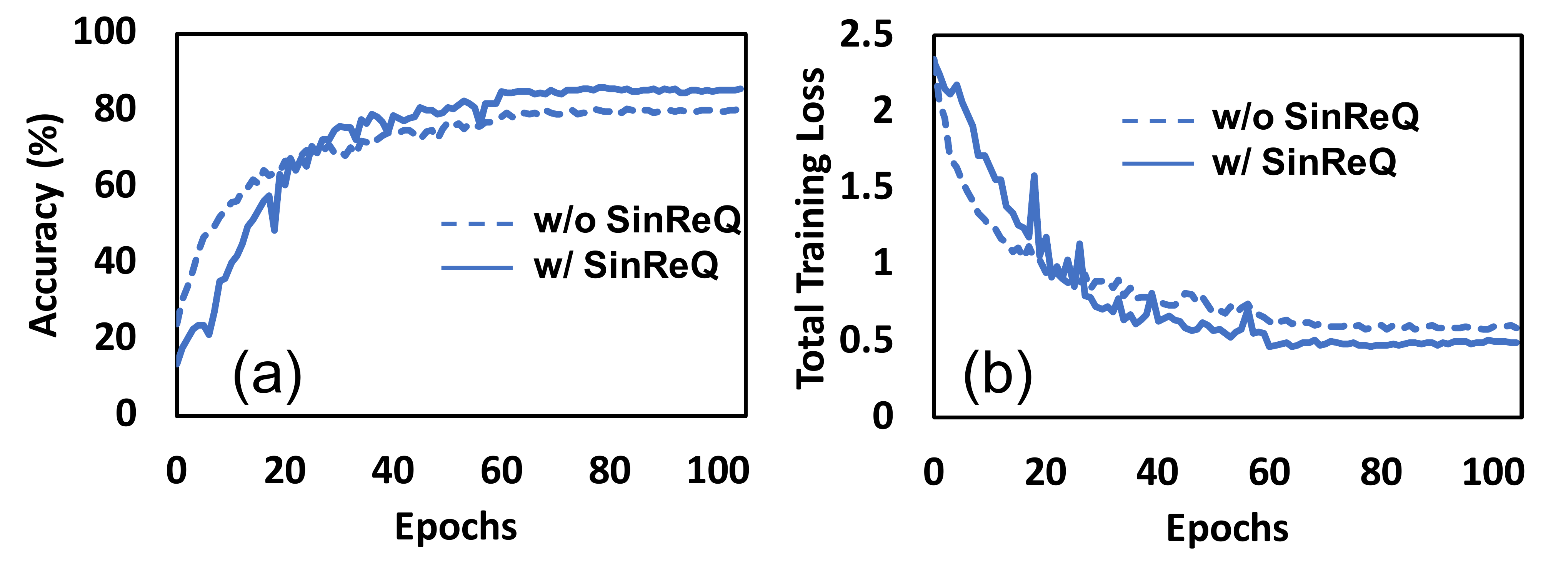}
  \caption{Comparing convergence behavior with and without \sinreq during training from scratch (a) accuracy, (b) training loss. Network: VGG-11, 2-bit DoReFa quantization}
  \label{fig:train}
\end{figure}

\begin{figure*}[t]
  \centering
	\includegraphics[width=0.6\linewidth]{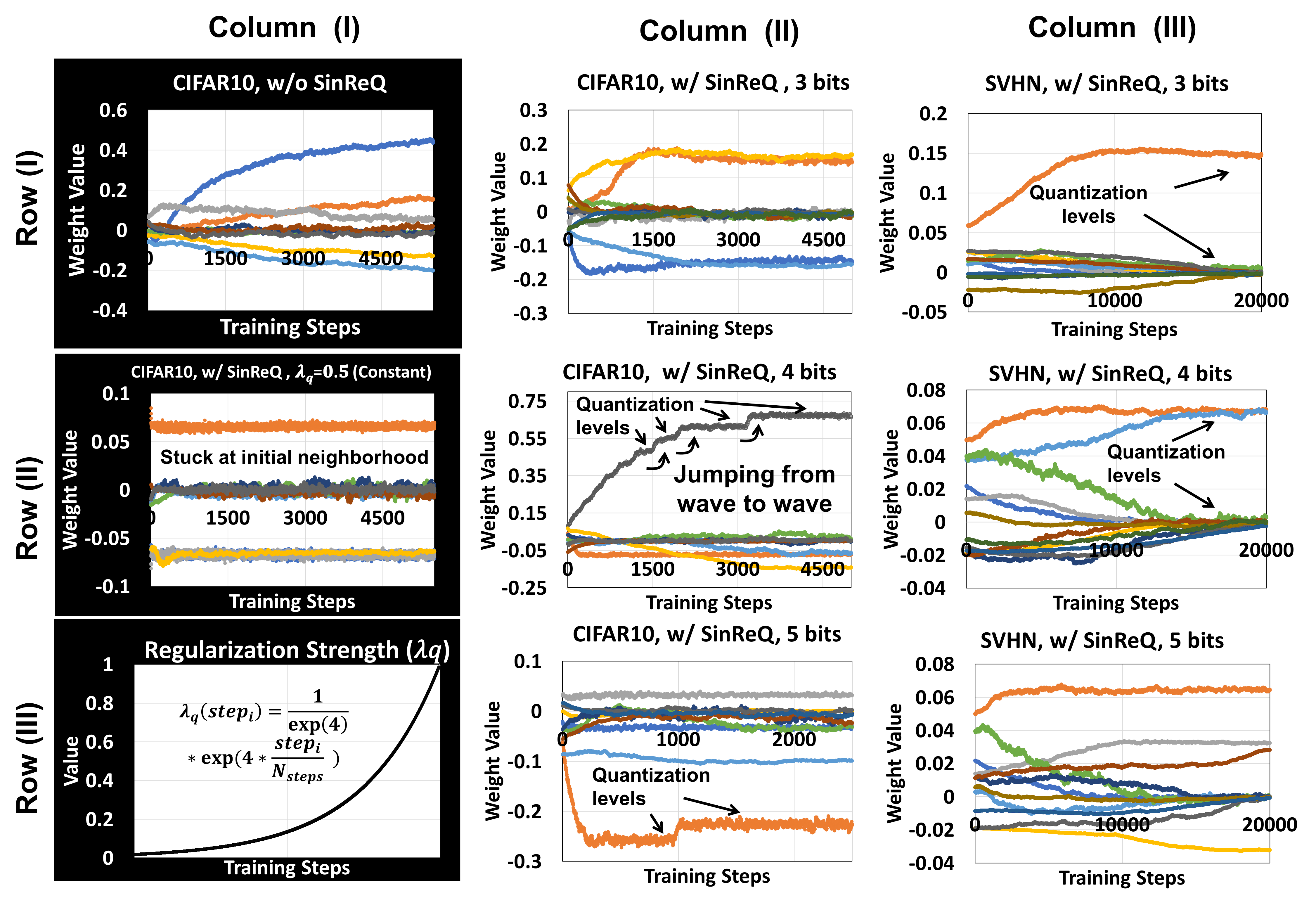}	
	\caption{Weight trajectories. The 10 colored lines in each plot denote the trajectory of 10 different weights.}
  \label{fig:w_traj}
\end{figure*}
\begin{figure}
  \centering
	\includegraphics[width=0.8\linewidth]{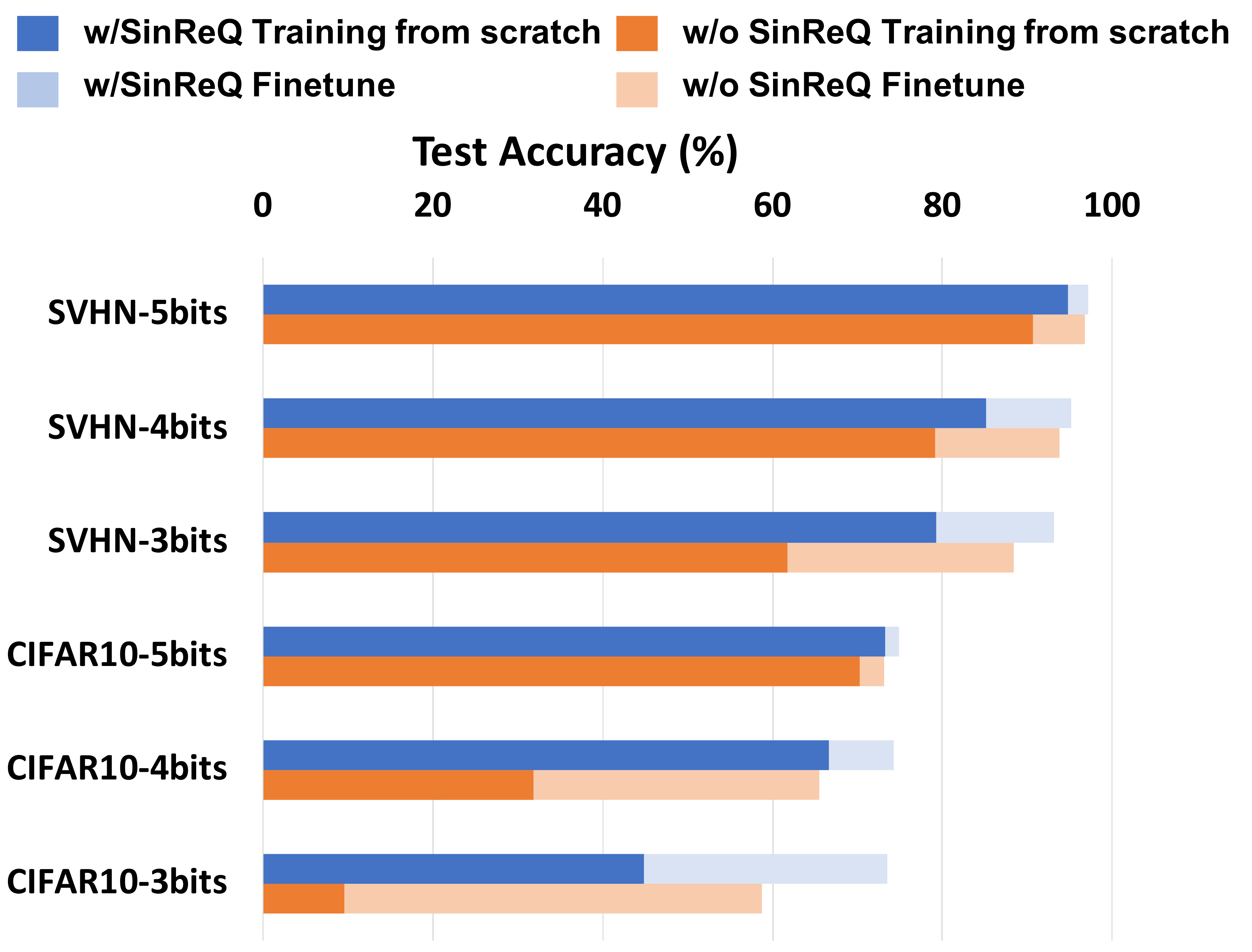}	
	\caption{Validation top-1 accuracy comparison w/o vs w/ SinReQ, for training from scratch and then fine-tuning.}
	\label{fig:acc_res}
\end{figure}

\niparagraph{Convergence analysis.}
Figure \ref{fig:finetune} shows the convergence behavior of SinReQ by visualizing both accuracy and regularization loss over finetuning epochs for two networks: CIFAR10 and SVHN.
As can be seen, the regularization loss (SinReQ Loss) is minimized across the finetuning epochs while the accuracy is maximized. This demonstrates a validity for the proposed regularization being able to optimize the two objectives simultaneously.
Figure \ref{fig:train} contrasts the convergence behavior with and without SinReQ for the case of training from scratch for VGG-11.  
As can be seen, at the onset of training, the accuracy in the presence of SinReQ is behind that without SinReQ. This can be explained as a result of optimizing for an extra objective in case of with SinReQ as compared to without.
Shortly thereafter, the regularization effect kicks in and eventually achieves $\sim6\%$  accuracy improvement.

%
The convergence behavior, however, is primarily controlled by the regularization strength $(\lambda_{q})$.
%
As briefly mentioned in section \ref{sec:sinreq}, $\lambda_q  \in  \lbrack 0, \infty) \,$ is a hyperparameter that weights the relative contribution of the proposed regularization objective to the standard accuracy objective.

%
In the context of neural networks, it is sometimes desirable to use a separate setting of $\lambda_{q}$ for each layer of the network. Since searching for the correct setting of muliple hyperparameters could be expensive process, it is still reasonable to use the same setting at all layers just to reduce the size of search space~\cite{Goodfellow-et-al-2016}.
Throughout our experiments, $\lambda_{q}$ is set the same across all layers and in the range of $0.5-10$. 
We reckon that careful setting of $\lambda_{q}$ across the layers and during the training epochs is essential for optimum results~\cite{DBLP:journals/corr/abs-1809-00095}.

\section{Intuition and Justification of Using SinReQ}

	\textbf{(1) Intuitively}: SinReQ regularization imposes a penalty proportional to the quantization error by aligning the minima of the sin regularizer with the quantization levels.
As such, it naturally minimizes the quantization error over the course of standard training process that uses gradient decent. 
%
%
Since this happens naturally, the accuracy loss, after direct quantization, is minimal. 
The SinReQ hyper-parameter (Regularization strength: $\lambda_q$) balances the tradeoff between the two objectives (the original loss and the quantization error) to ensure joint optimization of both objectives.
Additionally, SinReQ may also be thought of as a way to improve the generalization performance in the quantized domain where the derivative of SinReQ (proportional to the quantization error) acts as an additive noise component to the weight updates from the derivative of the empirical error during the training iterations.

\textbf{(2) Empirically:} 
We conduct an experiment that uses SinReQ \emph{for training from scratch---different from the experiments in the submission}.
Figure~\ref{fig:w_traj}-Row(I)-Column(I) shows weight trajectories without SinReQ as a point of reference.
Row(II)-Column(I) shows the weight trajectories when SinReQ is used with a constant $\lambda_q$.
As Figure~\ref{fig:w_traj}-Row(II)-Column(I) illustrates, using a constant $\lambda_q$ results in the weights being stuck in a region close to their initialization, (i.e., quantization objective dominates the accuracy objective), \emph{as pointed out by Reviewer 3}.
However, if we dynamically change the $\lambda_q$ following the exponential curve in Figure~\ref{fig:w_traj}-Row(III)-Column(I)) during the from-scratch training, the weights no longer get stuck.
Instead, the weights traverse the space (\emph{i.e., jump from wave to wave}) as illustrated in Figure~\ref{fig:w_traj}-Columns(II) and (III) for CIFAR and SVHN, respectively.
In these two columns, Rows (I), (II), (III), correspond to quantization with 3, 4, 5 bits, respectively.
Initially, the smaller $\lambda_q$ values allow the gradient descent to explore the optimization surface freely, as the training process moves forward, the larger $\lambda_q$ gradually engages the sine regularizer, and eventually pushes the weights close to the quantization levels.
As the results in Figure~\ref{fig:acc_res} show, training from scratch with SinReQ and dynamic $\lambda_q$, achieves strictly better accuracy than the baseline training without SinReQ across all cases (closing the accuracy gap to recover by 40.2\% on average). 
As shown, if the fine-tuning step is also engaged, the results improve further and in all cases SinReQ provides better accuracy.

\section{Conclusion}
Deep quantization of DNNs promises to be a powerful technique in reducing their complexity.
However, it comes with the vice of loss in accuracy that needs to be remedied.

This paper provided a new approach in using sinusoidal regularizations terms to push the weight values closer to the  quantized levels.
This mathematical approach is versatile and augments other quantized training algorithms by improving the quality of the network they train.
While this technique consistently improves the accuracy, \sinreq does not require changes to the base training algorithm or the neural network topology.

%
%

\section*{Acknowledgement}
This work was in part supported by Semiconductor Research Corporation contract \#2019-SD-2884, NSF awards CNS\#1703812, ECCS\#1609823, Air Force Office of Scientific Research (AFOSR) Young Investigator Program (YIP) award \#FA9550-17-1-0274, and gifts from Google, Microsoft, Xilinx, Qualcomm.

{\small
\bibliographystyle{reference/ieee}
\bibliography{reference/paper}
}

\end{document}